\documentclass[10pt,twocolumn,letterpaper]{article}
\usepackage[pagenumbers]{cvpr}

\usepackage{graphicx}
\usepackage{caption}
\usepackage{amsmath,amsfonts}
\usepackage{booktabs}
\usepackage{multirow}
\usepackage[dvipsnames]{xcolor}
\usepackage{quoting}
\usepackage{makecell}
\usepackage{CJK}
\usepackage{amssymb}
\definecolor{cvprblue}{rgb}{0.21,0.49,0.74}
\usepackage[pagebackref,breaklinks,colorlinks,allcolors=cvprblue]{hyperref}

\def\paperID{7869}
\def\confName{CVPR}
\def\confYear{2025}

\title{Sampling Innovation-Based Adaptive Compressive Sensing}

\author{Zhifu Tian, Tao Hu\thanks{Corresponding authors.} , Chaoyang Niu, Di Wu, Shu Wang\\
Information Engineering University, Zhengzhou, China\\
{\tt\small tzhifu@qq.com, hutaoengineering@163.com, niucy2017@outlook.com,} \\
{\tt\small wudipaper@sina.com, shu1008@mail.ustc.edu.cn}
}

\begin{document}
\maketitle

\begin{abstract}
	
	Scene-aware Adaptive Compressive Sensing (ACS) has attracted significant interest due to its promising capability for efficient and high-fidelity acquisition of scene images. ACS typically prescribes adaptive sampling allocation (ASA) based on previous samples in the absence of ground truth. However, when confronting unknown scenes, existing ACS methods often lack accurate judgment and robust feedback mechanisms for ASA, thus limiting the high-fidelity sensing of the scene. In this paper, we introduce a Sampling Innovation-Based ACS (SIB-ACS) method that can effectively identify and allocate sampling to challenging image reconstruction areas, culminating in high-fidelity image reconstruction. An innovation criterion is proposed to judge ASA by predicting the decrease in image reconstruction error attributable to sampling increments, thereby directing more samples towards regions where the reconstruction error diminishes significantly. A sampling innovation-guided multi-stage adaptive sampling (AS) framework is proposed, which iteratively refines the ASA through a multi-stage feedback process. For image reconstruction, we propose a Principal Component Compressed Domain Network (PCCD-Net), which efficiently and faithfully reconstructs images under AS scenarios. Extensive experiments demonstrate that the proposed SIB-ACS method significantly outperforms the state-of-the-art methods in terms of image reconstruction fidelity and visual effects. Codes are available at \url{https://github.com/giant-pandada/SIB-ACS_CVPR2025}.
	
\end{abstract}

\section{Introduction}
\label{sec:intro}
\begin{figure}
	\centering
	\vspace{-0.3cm}
	\setlength{\abovecaptionskip}{0.cm}
	\includegraphics[width=\columnwidth]{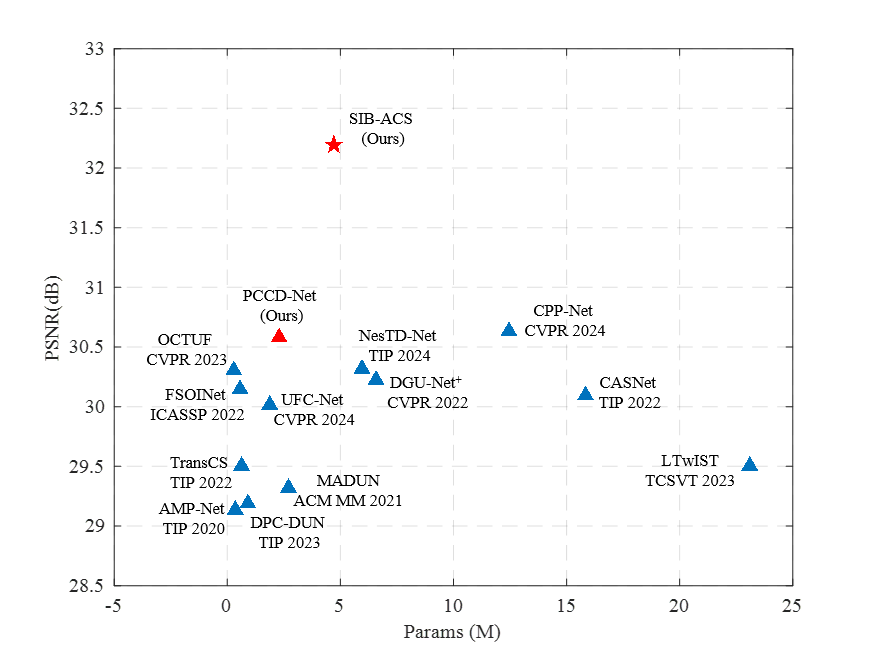}
	\caption{Comparison of average PSNR performance at the sampling rates of 0.10 and 0.25 on the BSD68~\cite{BSD68} and Urban100~\cite{Urban100} datasets between the proposed ACS model SIB-ACS, UCS model PCCD-Net and the state-of-the-art CS methods.}
	\label{fig:0}
	\vspace{-0.5cm}
\end{figure}

Compressive Sensing (CS) leverages the sparsity of signals, utilizing under-sampled bases for measurements to reconstruct the original signals with maximum fidelity. This technique has found broad application in various fields, including terahertz imaging~\cite{THz}, hyperspectral imaging~\cite{HI}, Magnetic Resonance Imaging (MRI)~\cite{MRI1}, and underwater imaging~\cite{UI}. The Uniform Compressive Sensing (UCS)~\cite{CATNet, SAUNet} model employs a uniform sampling rate across each image block for sample acquisition, followed by image reconstruction. However, due to the regional complexity variations often found in real-world images, areas of higher complexity require more extensive sampling for effective recovery. Adaptive Compressive Sensing (ACS)~\cite{DeepBCS, MB-RACS, MGSA, adasense} addresses this by dynamically adjusting the sampling allocation based on the content of individual image blocks, thereby enhancing the efficiency of scene sensing. ACS has been widely adopted in areas such as single-pixel imaging~\cite{SPI,SPI3}, image encryption~\cite{IE,IE2}, video compressive sensing~\cite{VCS,VCS2}, and medical imaging~\cite{MI}.

The most significant challenge for ACS is to perform adaptive sampling allocation (ASA) across various complex image regions, particularly in the absence of ground truth. When the ground truth is unavailable, existing ACS frameworks typically employ two-stage~\cite{RACSNet,Uformer-ICS} or multi-stage~\cite{Adacs, MB-RACS, MGSA, adasense} sampling, utilizing the sampling information from all previous stages. Multi-stage sampling holds immense potential as it iteratively reduces ASA errors through feedback mechanisms. When determining ASA, existing ACS methods analyze the previous measurements or the content of reconstructed images, allocating more samples to regions with larger measurement errors~\cite{Adacs,RACSNet,ME2}, higher cosine similarity in measurements~\cite{ME}, higher measurement scores~\cite{MGSA}, or more complex content in reconstructed images~\cite{Sa,texture}. However, existing ACS methods carry a potential risk of ASA errors. For methods in the measurement domain, the under-sampled measurement data are intrinsically more ill-posed than image data, leading to a significant potential for ASA errors. For methods in the image domain, the associated analysis indicators form a positive feedback loop with the sampling, which tends to concentrate sampling to the area of initial allocation~\cite{Adacs}, ultimately constraining the imaging capability of ACS.

To address the aforementioned issues, this paper proposes a Sampling Innovation-Based Adaptive Compressive Sensing (SIB-ACS) framework, which utilizes incremental image information from historical sampling and sampling increments to determine ASA through a multi-stage negative feedback process. Specifically, the proposed framework allocates sampling resources based on the size of the recovery of the principal components in the sampling increment, which we refer to as sampling innovation (SI). Furthermore, to fully harness SI in guiding adaptive sampling (AS), we have developed a multi-stage SI-guided AS model that corrects the ASA errors stage by stage in a negative feedback manner. For image reconstruction, we have designed a computationally efficient Principal Component Compressed Domain Network (PCCD-Net) to reconstruct high-fidelity images from adaptive sampling values. The proposed PCCD-Net employs proximal gradient descent (PGD) operations in the principal component (PC) image to compress the dimensions of PGD operations in the feature domain (FD), and simultaneously uses the PGD in the compressed FD to supplement the details of the features in the results of PGD in the PC image. As illustrated in Fig.~\ref{fig:0}, the proposed SIB-ACS adaptively senses the scenario, significantly outperforming the state-of-the-art CS methods. The main contributions are summarized as follows:
\begin{itemize}
	\item
	We propose a SIB-ACS framework that utilizes incremental image information from historical sampling and sampling increments to determine ASA, thereby enhancing the accuracy of ASA and achieving high-fidelity image reconstruction.
	\item
	We design a SI-guided multi-stage ASA model, which accomplishes multi-stage ASA at any sampling rate with a single model, facilitated by an image domain negative feedback mechanism.
	\item
	We introduce a PCCD-Net for image reconstruction in ACS scenarios, which leverages PGD operations in the PC image to compress the dimensions of PGD in the FD features, thereby significantly reducing the computational cost while maintaining high image reconstruction performance.
	\item
	Extensive experiments demonstrate that the proposed SIB-ACS significantly outperforms the state-of-the-art CS methods.
\end{itemize}

\section{Related work}
\label{sec:related work}

\textbf{Adaptive Compressive Sensing Methods}. ACS dynamically allocates sampling rates based on the complexity of block image content, thereby enhancing image sensing capabilities. On the one hand, when scene images are available, existing studies have achieved significant results by guiding AS based on scene content analysis indicators such as texture~\cite{texture,gradient}, edge~\cite{edge,edge2,THz}, entropy~\cite{e-s-v}, variance~\cite{VCS,VCS2}, gradient~\cite{gradient}, wavelet components~\cite{SPI,wavelet1,wavelet2,wavelet3,HI}, energy~\cite{UI,UI2,energy1}, or saliency distribution~\cite{Sa,CASNet}. On the other hand, when scenes are unknown, existing studies employ two-stage~\cite{RACSNet,Uformer-ICS} or multi-stage models~\cite{Adacs, MB-RACS, MGSA, adasense} for AS, using the sampling from all previous AS stages to make ASA judgments. Generally, ASA methods include measurement domain (MD) and image domain (ID) methods. Specifically, existing MD methods use measurement value-related indicators for ASA, allocating more samples to areas with high indicators such as scene measurement error~\cite{Adacs, RACSNet,ME2}, cosine similarity in measurements~\cite{ME}, and network scores of measurements~\cite{MGSA}. In addition, ID methods~\cite{Uformer-ICS} first reconstruct all previous sampling values into preliminary images, then allocate more samples to areas with high analysis indicators in the content of the preliminary reconstructed images. These ASA methods have achieved significant results by distinguishing image regions using different strategies~\cite{Adacs,Uformer-ICS}.

However, when the scene is unknown, the ASA results of existing ACS methods fall short of the ideal effect. MD methods tend to have noticeable judgment errors due to the inherent ill-posedness of under-sampled measurement data in comparison to image data. In ID methods, the sampling and image analysis indicators form a positive feedback loop, making it difficult to correct ASA errors in the multi-stage ACS framework~\cite{Adacs}, resulting in its accuracy heavily relying on the initial reconstructed image quality, which often degrades significantly under low sampling conditions. These limitations in existing ASA methods have hindered the imaging performance of ACS to varying degrees in terms of ASA accuracy, obstructing the practical application of high-fidelity ACS. For unknown scenes, we propose a SIB-ACS method that utilizes all historical sampling and sampling increment information to implement an image-level negative feedback multi-stage AS framework.

\noindent\textbf{CS Reconstruction Networks}. Leveraging the success of machine vision~\cite{gao1,gao2}, image CS has developed purely deep learning-based (purely DL) networks and deep unfolding networks (DUNs). Purely DL methods utilize the mapping capabilities of neural networks to learn the inverse problem of image reconstruction, resulting in pure CNN networks~\cite{CSNet, DR2-Net, SCSNet, AutoBCS, MR-CCSNet, SCS-GNet}, attention-based networks~\cite{TCSNet}, and hybrid networks of CNN and Transformer~\cite{NL-CSNet, CSformer, DPA, MR-CCSNet,MCFDNet,s2cs} for CS. Purely DL networks improve the capabilities for block artifact removal and image denoising by harnessing the networks' mapping abilities but suffer from the infamous black box problem. On the other hand, DUNs integrate traditional optimization algorithms with neural networks, enhancing image reconstruction performance by using the networks' robust mapping capabilities and injecting measurement physics information~\cite{FICS}. Researchers have extensively explored the networks that unfold various traditional optimization algorithms~\cite{ADMM-CSNet,CATNet,SAUNet,ISTA-Net,iPiano-Net,Neumann,G2DUN,SCNet}. Notably, it is proven that large dimensions of physics measurement injection~\cite{HiTDUN,IDM,PRL,Uformer-ICS,FHDUN,D3U,D3C2} and information flow~\cite{MAPUN,DGUNet,SODAS-Net,MACNet,DUN-CSNet,GCDUN} can further enhance DUNs' performance. However, a large physics injection dimension usually implies higher computational cost, such as for gradient descent (GD) operations. Particularly compared to UCS, ACS requires more sampling in complex areas, leading to a larger physics injection computational costs due to the increased sampling matrix. In this paper, we introduce a PCCD-Net for ACS, which uses the principal component GD to compress the dimensionality of the feature domain GD, thereby achieving high image fidelity with fewer operations.
\begin{figure*}[ht]
	\centering
	\setlength{\abovecaptionskip}{-0.3cm}
	\includegraphics[width=.8\linewidth]{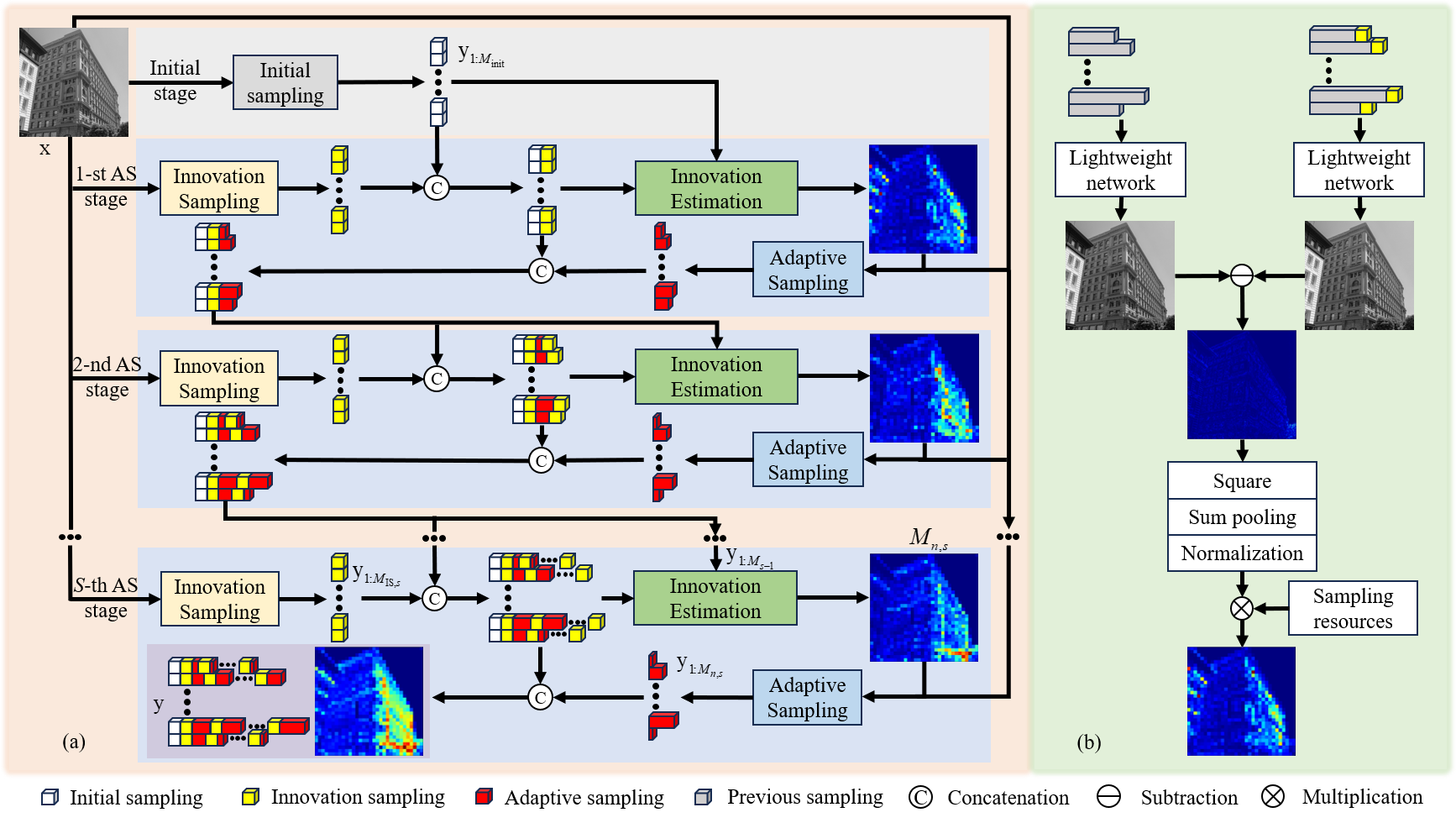} \rule{.8\linewidth}{0pt}
	\caption{The overview of the proposed sampling innovation-based ASM. (a) Innovation-guided multi-stage AS, (b) Innovation Estimation (IE) based on the reconstructed image information from sampling values before and after Innovation Sampling (IS).}
	\label{fig:1}
	\vspace{-0.5cm}
\end{figure*}

\section{Proposed method}
\label{sec:proposed method}

\begin{figure*}
	\centering
	\setlength{\abovecaptionskip}{-0.2cm}
	\includegraphics[width=.8\linewidth]{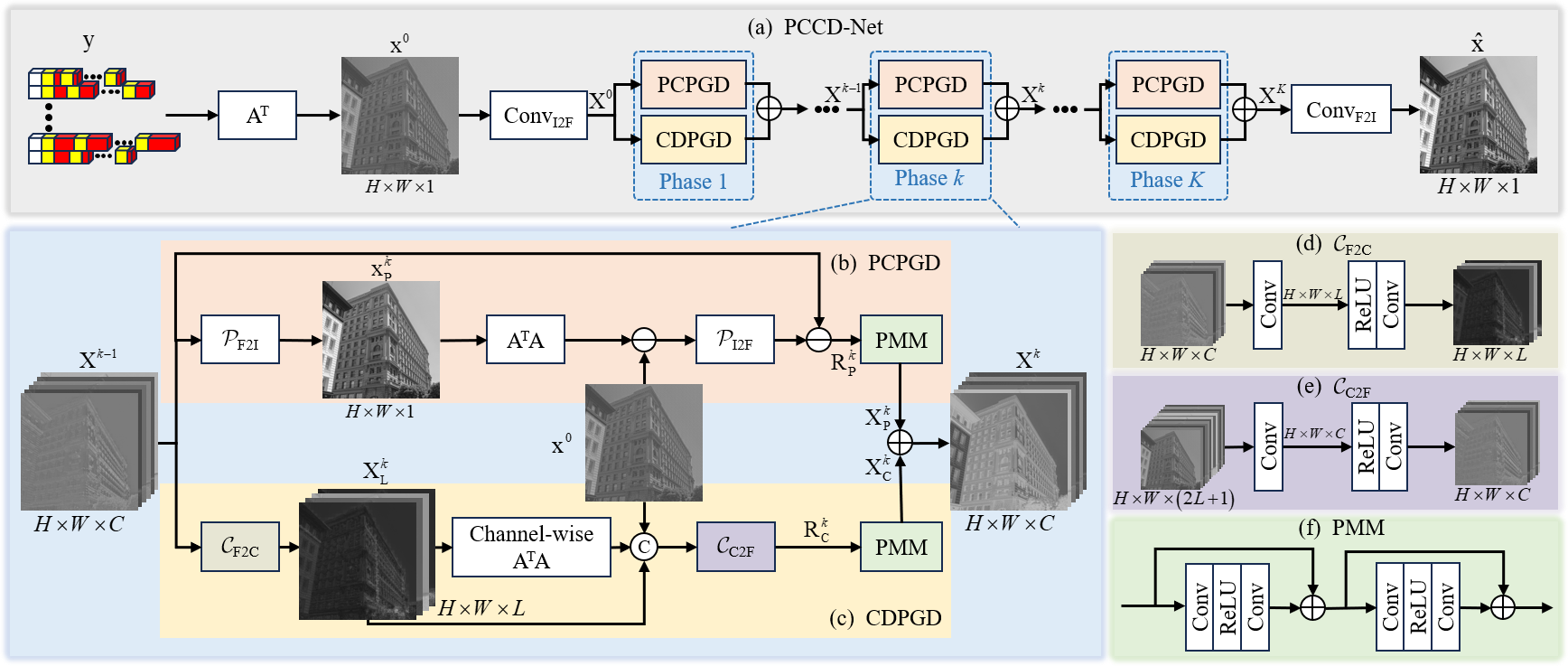} \rule{.8\linewidth}{0pt}
	\caption{The overview of the proposed PCCD-Net for image reconstruction. (a) Deep reconstruction process, (b) PCPGD path, (c) CDPGD path, (d) Convolutional block that transitions features from the FD to the CD, (e) Convolutional block that transitions features from the CD back to the FD, (f) Proximal Mapping Module (PMM).}
	\label{fig:2}
	\vspace{-0.5cm}
\end{figure*}

In this section, we provide a detailed introduction to the proposed SIB-ACS, which includes an adaptive sampling module (ASM) and an image reconstruction module, as depicted in Figs.~\ref{fig:1} and~\ref{fig:2}, respectively.

\subsection{ASM}
\label{sec:asm}
\textbf{Innovation Criterion}. ACS optimizes the sampling strategy with the objective function of minimizing the error in the reconstructed image. Accordingly, when the ground truth is unavailable, we design an innovation method to allocate the sampling resources based on the image successfully reconstructed from historical sampling values. Specifically, innovation refers to the increase in the components of the reconstructed image or the decrease in the reconstruction error that results from an increase in the sampling rate. The sampling used to measure the innovation is referred to as innovation sampling. In detail, innovation can be defined as
\begin{equation}
	\alpha = \parallel\!{\hat{\mathbf{x}}_{\mathrm{IS}}-\hat{\mathbf{x}}_{\mathrm{HM}}}\!\parallel_2^2 = \parallel\!{(\mathbf{x}-\hat{\mathbf{x}}_{\mathrm{HM}})-(\mathbf{x}-\hat{\mathbf{x}}_{\mathrm{IS}})}\!\parallel_2^2,
	\label{eq:1}
\end{equation}
where $\hat{\mathbf{x}}_{\mathrm{HM}}$ represents the image reconstructed from the historical measurements, $\hat{\mathbf{x}}_{\mathrm{IS}}$ denotes the image reconstructed after innovation sampling, and $\mathbf{x}$ is the original image. The innovation method adaptively determines the number of samples for each image block, based on the magnitude of the innovation within the same image. With this method, the adaptively allocated number of samples can be expressed as
\begin{equation}
	M_n = M_{\mathrm{ASR}}\frac{\parallel\!\alpha_n\!\parallel_2^2}{\sum\limits_{n}\parallel\!\alpha_n\!\parallel_2^2},
	\label{eq:2}
\end{equation}
where $M_{\mathrm{ASR}}$ denotes the adaptive sampling resources, while $\alpha_n$ and $M_n$ represent the innovation and the adaptively allocated sample count for the $n$-th image block, respectively. As illustrated by Eq.~(\ref{eq:1}), innovation is a direct estimation of the decrease in reconstruction error. Therefore, the direction guided by innovation-based AS aligns with the goal of reducing the image reconstruction error, ensuring that the AS progresses towards minimizing the total image reconstruction error. Furthermore, innovation is a relative measure, not an absolute one. As the number of samples guided by innovation within an image block increases, the innovation will gradually decrease once the principal components of the image are restored, which creates a negative feedback mechanism, guiding AS towards a larger recovery of innovation, thereby enabling the possibility of multi-stage AS at the image level.

\noindent\textbf{Innovation-guided ASM}. The innovation-guided multi-stage ASM comprises two stages: the initial sampling stage and the multi-stage AS stage, as illustrated in Fig.~\ref{fig:1}. 

The initial sampling stage acts as the startup stage of the ASM. For all image blocks, a uniform initial sampling count $M_{\mathrm{init}}$ or sampling rate $SR_{\mathrm{init}}$ is employed to acquire the initial sampling values. Therefore, the initial sampling stage can be expressed as 
\begin{equation}
	\mathbf{y}_{1:M_{\mathrm{init}}} = \mathbf{A}_{1:M_{\mathrm{init}}}\mathbf{x},
	\label{eq:3}
\end{equation}
where $\mathbf{A}_{1:M_{\mathrm{init}}}$ and $\mathbf{y}_{1:M_{\mathrm{init}}}$ respectively denote the initial sampling matrix and the initial sampling values with a sample count of $M_{\mathrm{init}}$.

The multi-stage AS stage effectively leverages the multi-stage negative feedback mechanism to progressively eliminate the residual innovation in each block image. Specifically, each AS stage involves three steps: Innovation Sampling (IS), Innovation Estimation (IE), and Adaptive Sampling (AS), as depicted in Fig.~\ref{fig:1}(a). Particularly, the first step, IS, is based on the sampling distribution at the $s-1$-th stage, conducting uniform sampling with a sample count of $M_{\mathrm{IS}}$ or a sampling rate of $SR_{\mathrm{IS}}$, to uniformly probe the innovation distribution. Therefore, IS at the $s$-th stage can be represented as
\begin{equation}
	\mathbf{y}_{1:M_{\mathrm{IS},s}} = \mathbf{A}_{1:M_{\mathrm{IS},s}}\mathbf{x},
	\label{eq:4}
\end{equation}
where $\mathbf{A}_{1:M_{\mathrm{IS},s}}$ and $\mathbf{y}_{1:M_{\mathrm{IS},s}}$ respectively denote the IS matrix and the IS values with a sample count of $M_{\mathrm{IS}}$ at the $s$-th stage. The sampling values $\mathbf{y}_{1:M_{s-1}+M_{\mathrm{IS},s}}$ after the IS at the $s$-th stage are formed by concatenating the sampling value of pre-s-1 stage $\mathbf{y}_{1:M_{s-1}}$ and $\mathbf{y}_{1:M_{\mathrm{IS},s}}$, and the sampling matrix $\mathbf{A}_{1:M_{s-1}+M_{\mathrm{IS},s}}$ is formed by concatenating $\mathbf{A}_{1:M_{s-1}}$ and $\mathbf{A}_{1:M_{\mathrm{IS},s}}$. 

Subsequently, as shown in Fig.~\ref{fig:1}(b), the second step, IE at the $s$-th stage, initially employs lightweight networks to swiftly reconstruct the images $\hat{\mathbf{x}}_{s-1}$ and $\hat{\mathbf{x}}_{\mathrm{IS},s}$ before and after the IS, respectively, 
\begin{equation}
	\hat{\mathbf{x}}_{s-1} = \mathcal{F}_{s_1}(\mathbf{y}_{1:M_{s-1}}, \mathbf{A}_{1:M_{s-1}}),
	\label{eq:5}
\end{equation}
\begin{equation}
	\hat{\mathbf{x}}_{\mathrm{SI},s} = \mathcal{F}_{s_2}(\mathbf{y}_{1:M_{s-1}+M_{\mathrm{IS},s}}, \mathbf{A}_{1:M_{s-1}+M_{\mathrm{IS},s}}),
	\label{eq:6}
\end{equation}
where $\mathcal{F}_{s_1}$ and $\mathcal{F}_{s_2}$ respectively denote the lightweight reconstruction networks before and after IS. Subsequently, the innovation at the $s$-th stage can be computed according to Eq.~(\ref{eq:1}),
\begin{equation}
	\alpha_s = \parallel\!{\hat{\mathbf{x}}_{\mathrm{IS},s}-\hat{\mathbf{x}}_{s-1}}\!\parallel_2^2.
	\label{eq:7}
\end{equation}

Next, the innovation across the entire image is normalized to derive the weight $\frac{\parallel\!\alpha_n\!\parallel_2^2}{\sum\limits_{n}\parallel\!\alpha_n\!\parallel_2^2}$ for block image sampling allocation. Additionally, in the multi-stage AS process, an image with height $H$ and width $W$ respectively, is divided into $N$ block images of size $B$. The total sampling resource that can be allocated at each AS stage of the $S$-stage ASM, given a total sampling rate of $SR$, is computed as $M_{\mathrm{ASR},s}=\lceil \frac{H}{B} \rceil \lceil \frac{W}{B} \rceil B^2(\frac{SR-SR_{\mathrm{init}}}{S}-SR_{\mathrm{IS}})$, where $\lceil\rceil$ denotes the operation of taking the ceiling. Therefore, the number of adaptive samples for the $n$-th image block at the $s$-th stage is
\begin{equation}
	M_{n,s} = M_{\mathrm{ASR},s}\frac{\parallel\!\alpha_{n,s}\!\parallel_2^2}{\sum\limits_{n}\parallel\!\alpha_{n,s}\!\parallel_2^2}.
	\label{eq:8}
\end{equation}

In the third step, AS is applied to the image, utilizing the adaptive sampling count for the $n$-th image block in the $s$-th stage, as determined in the second step. The maximum full sampling rate is assumed to be $1$ by default, so we set the maximum sampling rate of each image block after the AS in the $s$-th stage to be $s/S$. Upon completion of the $s$-th stage AS, the sampling values $\mathbf{y}_{1:M_s}$ from the preceding $s$ stages are obtained by concatenating $\mathbf{y}_{1:M_{s-1}+M_{\mathrm{IS},s}}$ and $\mathbf{y}_{1:M_{n,s}}$.

This AS process is iterated across $S$ stages, ultimately yielding the final sampling value $y$ from the ASM.
\subsection{PCCD-Net for image reconstruction}
\label{sec:PCCD}
The Principal Component Compressed Domain Network (PCCD-Net) is depicted in Fig.~\ref{fig:2}. Firstly, the initial reconstructed image is generated through the deconvolution of the adaptive sampling values and the corresponding sampling matrix,
\begin{equation}
	\mathbf{x}^0 = \mathbf{A}\textsuperscript{T}\mathbf{y},
	\label{eq:9}
\end{equation}
where $\mathbf{A}\textsuperscript{T}$ represents the transpose of $\mathbf{A}$. Subsequently, the initial reconstructed image is transformed into initial features with $C$ channels,
\begin{equation}
	\mathbf{X}^0 = \mathrm{Conv_{I2F}}(\mathbf{x}^0),
	\label{eq:10}
\end{equation}
where $\mathrm{Conv_{I2F}}$ denotes the convolution operation that transforms the image to features. Next, the initial features are updated to the final reconstructed features through $K$ phases of Proximal Gradient Descent (PGD)~\cite{PGD} iteration. Specifically, each iterative stage is divided into a Principal Component PGD (PCPGD) path and a Compressed Domain PGD (CDPGD) path, as shown in Fig.~\ref{fig:2}(b) and Fig.~\ref{fig:2}(c), respectively. The PCPGD first aggregates the reconstructed features from the previous stage into a principal component (PC) image through a convolution operation. Subsequently, the PCPGD calculates the gradient of the PC image and uses a convolution operation to extend the gradient image back into the feature domain (FD), where the proximal mapping operation is executed. The PCPGD process can be represented as follows:
\begin{equation}
	\mathbf{x}^{k}_{\mathrm{p}} = \mathcal{P}_{\mathrm{F2I}}(\mathbf{X}^{k-1}),
	\label{eq:11}
\end{equation}
\begin{equation}
	\mathbf{R}^{k}_{\mathrm{p}} = \mathbf{X}^{k-1}-\mathcal{P}_{\mathrm{I2F}}(\mathbf{A}\textsuperscript{T}(\mathbf{A}\mathbf{x}^{k}_{p}-\mathbf{y})),
	\label{eq:12}
\end{equation}
\begin{equation}
	\mathbf{X}^{k}_{\mathrm{p}} = \mathcal{F}_{\mathrm{PM}}(\mathbf{R}^{k}_{\mathrm{p}}),
	\label{eq:13}
\end{equation}
where $\mathbf{x}^{k}_{\mathrm{p}}$ and $\mathbf{X}^{k}_{\mathrm{p}}$ respectively represent the PC image and features at the $k$-th phase, while $\mathcal{P}_{\mathrm{I2F}}$ and $\mathcal{P}_{\mathrm{F2I}}$ denote the convolution operations that transition from the PC image to features and from features back to the PC image, respectively. $\mathcal{F}_{\mathrm{PM}}$ stands for the proximal mapping module (PMM), as shown in Fig.~\ref{fig:2}(f). The CDPGD integrates the reconstructed features from the previous stage into compressed domain (CD) complementary features of dimension $L$ via a convolution block, and then computes the gradient in a channel-wise manner. Subsequently, the gradient descent process is completed by executing a convolution block operation following concatenating features of different dimensions. Finally, the CD gradient is restored to the FD of dimension $C$, where the PMM is implemented. The CDPGD process can be represented as:
\begin{equation}
	\mathbf{X}^{k}_{\mathrm{L}} = \mathcal{C}_{\mathrm{F2C}}(\mathbf{X}^{k-1}),
	\label{eq:14}
\end{equation}
\begin{equation}
	\mathbf{R}^{k}_{\mathrm{c}} = \mathcal{C}_{\mathrm{C2F}}(\mathrm{Concat}(\mathbf{X}^{k}_{L},\mathbf{A}\textsuperscript{T}\mathbf{A}\mathbf{X}^{k}_{L},\mathbf{A}\textsuperscript{T}\mathbf{y})),
	\label{eq:15}
\end{equation}
\begin{equation}
	\mathbf{X}^{k}_{\mathrm{c}} = \mathcal{F}_{\mathrm{PM}}(\mathbf{R}^{k}_{\mathrm{c}}),
	\label{eq:16}
\end{equation}
where $\mathbf{x}^{k}_{\mathrm{c}}$ and $\mathbf{X}^{k}_{\mathrm{L}}$ denote the CD image and features at the $k$-th phase, respectively. Meanwhile, $\mathcal{C}_{\mathrm{F2C}}$ and $\mathcal{C}_{\mathrm{C2F}}$ respectively represent the convolutional operations that transition features from the FD to the CD and from the CD back to the FD, as shown in Fig.~\ref{fig:2}(d) and Fig.~\ref{fig:2}(e), respectively. At this point, the image features updated by both PCPGD and CDPGD are amalgamated in a complementary fashion, culminating in the comprehensive image features $\mathbf{X}^{k}=\mathbf{X}^{k}_{\mathrm{p}}+\mathbf{X}^{k}_{\mathrm{c}}$. Ultimately, following $K$ phases of iterative update, the comprehensive image features are aggregated via a convolution operation to yield the reconstructed image,
\begin{equation}
	\hat{\mathbf{x}} = \mathrm{Conv_{F2I}}(\mathbf{X}^{K}),
	\label{eq:17}
\end{equation}
where $\mathrm{Conv_{F2I}}$ represents the convolution operation that transitions from features to the image.

\subsection{Loss Function}
\label{sec:Loss}
Considering the quality of both the reconstructed image pixels and textures in the SIB-ACS, we employ a combined loss function consisting of $l_1$ loss and $\mathrm{SSIM}$ loss~\cite{CSformer,Loss}, expressed as follows:
\begin{equation}
	\mathcal{L}(\Theta) = \mathcal{L}_{l_1}(\Theta)+ \mu\mathcal{L}_{\mathrm{SSIM}}(\Theta),
	\label{eq:13}
\end{equation}
where $\Theta$ represents all the learnable parameters of the model, $\mathcal{L}_{l_1}$ and $\mathcal{L}_{\mathrm{SSIM}}$ represent $l_1$ loss and $\mathrm{SSIM}$ loss, respectively, and $\mu$ is an empirical constant.
\begin{table*}[ht]
	\caption{Average PSNR ($\mathrm{dB}$) and SSIM results of recent uniform and adaptive CS methods on BSD68~\cite{BSD68} and Urban100~\cite{Urban100} with different sampling ratios. The best and second-best results are marked in red and blue colors, respectively.}
	%\small
	\centering
	\resizebox{0.8\linewidth}{!}{
		\begin{tabular}{c|c|l|cccccc}
			\toprule[1pt]
			\bottomrule[0.5pt]
			\rule{0pt}{10pt}\multirow{2}{*}{Dataset} & \multicolumn{2}{c|}{\multirow{2}{*}{Models}} &
			\multicolumn{6}{c}{CS Ratio}\cr\cline{4-9} & \multicolumn{2}{c|}{}
			& \rule{0pt}{10pt}$SR = 0.10$ & $SR = 0.25$ & $SR = 0.30$ & $SR = 0.40$ & $SR = 0.50$ & Average \cr
			\cline{1-9}
			\multirow{17}{*}{BSD68}
			&\multirow{13}{*}{\makecell[c]{Uniform \\ Sampling}}
			& \rule{0pt}{10pt}AMP-Net-9-BM (TIP 2021)~\cite{AMP-Net} & 27.86/0.7926 & 31.74/0.9048 & 32.84/0.9240 & 34.86/0.9508 & 36.82/0.9680 & 32.82/0.9080\cr
			& & MADUN (ACM MM 2021)~\cite{MADUN} & 26.83/0.7620 & 30.81/0.8845 & 31.87/0.9068 & 33.81/0.9376 & 35.76/0.9582 & 31.82/0.8898 \cr
			& & TransCS (TIP 2022)~\cite{TransCS} & 27.86/0.8086 & 31.74/0.9121 & 32.56/0.9276 & 34.82/0.9540 & 36.81/0.9699 & 32.76/0.9144 \cr
			& & FSOINet (ICASSP 2022)~\cite{FSOINET} & 28.27/0.8187 & 32.21/0.9183 & 33.29/0.9348 & 35.32/0.9577 & 37.34/0.9727 & 33.29/0.9204 \cr
			& & DGUNet$^+$ (CVPR 2022)~\cite{DGUNet} & 28.13/0.8165 & 31.97/0.9158 & -/- & -/- & 37.04/0.9718 & -/- \cr
			& & LTwIST (TCSVT 2023)~\cite{LTwSTA} & 27.85/0.8082 & 31.64/0.9102 & -/- & -/- & -/- & -/- \cr
			& & DPC-DUN (TIP 2023)~\cite{DPC-DUN} & 26.77/0.7608 & 30.70/0.8832 & 31.75/0.9050 & 33.70/0.9364 & 35.62/0.9574 & 31.71/0.8886 \cr
			& & OCTUF (CVPR 2023)~\cite{OCTUF} & 28.28/0.8177 & 32.24/0.9185 & 33.32/0.9348 & 35.35/0.9578 & 37.41/\textcolor{blue}{0.9729} & 33.32/0.9203 \cr
			& & NesTD-Net (TIP 2024)~\cite{NesTD-Net} & 28.28/\textcolor{blue}{0.8231} & 32.30/\textcolor{blue}{0.9203} & 33.22/0.9354 & 35.37/\textcolor{blue}{0.9586} & -/- & -/- \cr
			& & UFC-Net (CVPR 2024)~\cite{UFC} & 27.95/0.8086 & 31.74/0.9093 & -/- & -/- & -/- & -/- \cr
			& & CPP-Net (CVPR 2024)~\cite{CPP} & 28.41/0.8227 & 32.25/0.9188 & 33.34/0.9353 & 35.33/0.9575 & 37.30/0.9722 & 33.33/0.9213\cr
			\cline{2-9}
			&\multirow{4}{*}{\makecell[c]{Adaptive \\ Sampling}}
			&\rule{0pt}{10pt}CASNet (TIP 2022)~\cite{CASNet} & 28.41/0.8230 & 32.31/0.9196 & 33.39/\textcolor{blue}{0.9358} & 35.44/0.9581 & 37.49/0.9728 & 33.41/\textcolor{blue}{0.9219} \cr
			& & AMS-Net (TMM 2022)~\cite{AMS-Net} & \textcolor{blue}{29.36}/0.8073 & \textcolor{blue}{33.53}/0.9146 & \textcolor{blue}{34.67}/0.9321 & \textcolor{blue}{36.88}/0.9555 & \textcolor{blue}{39.20}/0.9703 & \textcolor{blue}{34.73}/0.9160 \cr
			& & Uformer-ICS (TSC 2023)~\cite{Uformer-ICS} & 27.75/0.7978 & 31.74/0.9006 & 32.62/0.9179 & 34.15/0.9425 & 35.49/0.9590 & 32.35/0.9036 \cr
			& & \textbf{SIB-ACS (ours)} & \textcolor{red}{29.54}/\textcolor{red}{0.8401} & \textcolor{red}{34.35}/\textcolor{red}{0.9312} & \textcolor{red}{35.72}/\textcolor{red}{0.9455} & \textcolor{red}{38.38}/\textcolor{red}{0.9653} & \textcolor{red}{41.14}/\textcolor{red}{0.9779} & \textcolor{red}{35.83}/\textcolor{red}{0.9320} \cr
			\cline{1-9}
			\multirow{17}{*}{Urban100}
			&\multirow{13}{*}{\makecell[c]{Uniform \\ Sampling}}
			& \rule{0pt}{10pt}AMP-Net-9-BM (TIP 2021)~\cite{AMP-Net} & 26.03/0.8148 & 30.87/0.9199 & 32.18/0.9363 & 34.35/0.9577 & 36.31/0.9710 & 31.95/0.9199 \cr
			& & MADUN (ACM MM 2021)~\cite{MADUN} & 27.11/0.8389 & 32.52/0.9344 & 33.74/0.9469 & 35.72/0.9630 & 37.70/0.9744 & 33.36/0.9315 \cr
			& & TransCS (TIP 2022)~\cite{TransCS} & 26.72/0.8411 & 31.70/0.9329 & 31.95/0.9382 & 35.20/0.9647 & 37.18/0.9760 & 32.55/0.9306\cr
			& & FSOINet (ICASSP 2022)~\cite{FSOINET} & 27.53/0.8624 & 32.60/0.9428 & 33.82/0.9538 & 35.90/0.9686 & 37.77/0.9775 & 33.52/0.9410 \cr
			& & DGUNet$^+$ (CVPR 2022)~\cite{DGUNet} & 28.01/0.8707 & 32.76/0.9450 & -/- & -/- & 37.63/0.9783 & -/- \cr
			& & LTwIST (TCSVT 2023)~\cite{LTwSTA} & 26.76/0.8461 & 31.78/0.9347 & -/- & -/- & -/- & -/- \cr
			& & DPC-DUN (TIP 2023)~\cite{DPC-DUN} & 26.95/0.8357 & 32.34/0.9319 & 33.50/0.9466 & 35.55/0.9620 & 37.48/0.9734 & 33.16/0.9299 \cr
			& & OCTUF (CVPR 2023)~\cite{OCTUF} & 27.76/0.8622 & 32.96/0.9444 & 34.18/0.9553 & 36.22/0.9696 & 38.23/0.9794 & 33.87/0.9422 \cr
			& & NesTD-Net (TIP 2024)~\cite{NesTD-Net} & 27.74/0.8673 & 32.94/0.9444 & 33.44/0.9513 & 35.87/0.9680 & -/- & -/- \cr
			& & UFC-Net (CVPR 2024)~\cite{UFC} & 27.54/0.8581 & 32.81/0.9421 & -/- & -/- & -/- & -/- \cr
			& & CPP-Net (CVPR 2024)~\cite{CPP} & 28.48/\textcolor{blue}{0.8798} & 33.37/\textcolor{blue}{0.9483} & 34.56/\textcolor{blue}{0.9583} & 36.48/\textcolor{blue}{0.9712} & 38.29/\textcolor{blue}{0.9799} & 34.24/\textcolor{blue}{0.9475}\cr
			\cline{2-9}
			&\multirow{4}{*}{\makecell[c]{Adaptive \\ Sampling}}
			&\rule{0pt}{10pt}CASNet (TIP 2022)~\cite{CASNet} & 27.43/0.8613 & 32.20/0.9394 & 33.36/0.9510 & 35.45/0.9667 & 37.43/0.9770 & 33.17/0.9391\cr
			& & AMS-Net (TMM 2022)~\cite{AMS-Net} & 28.04/0.8398 & 33.22/0.9326 & 34.35/0.9451 & 36.38/0.9614 & 38.33/0.9716 & 34.06/0.9301\cr
			& & Uformer-ICS (TSC 2023)~\cite{Uformer-ICS} & \textcolor{blue}{29.10}/0.8589 & \textcolor{blue}{33.87}/0.9353 & \textcolor{blue}{35.02}/0.9464 & \textcolor{blue}{37.11}/0.9617 & \textcolor{blue}{39.06}/0.9720 & \textcolor{blue}{34.83}/0.9349\cr
			& & \textbf{SIB-ACS (ours)} & \textcolor{red}{29.70}/\textcolor{red}{0.8859} & \textcolor{red}{35.15}/\textcolor{red}{0.9516} & \textcolor{red}{36.50}/\textcolor{red}{0.9605} & \textcolor{red}{38.93}/\textcolor{red}{0.9727} & \textcolor{red}{41.28}/\textcolor{red}{0.9807} & \textcolor{red}{36.31}/\textcolor{red}{0.9503} \cr
			\toprule[0.5pt]
			\toprule[1pt]
	\end{tabular}}
	\label{tab:1}
\end{table*}
\begin{figure*}
	\vspace{-0.4cm}
	\centering
	\setlength{\abovecaptionskip}{0.cm}
	\includegraphics[width=0.8\linewidth]{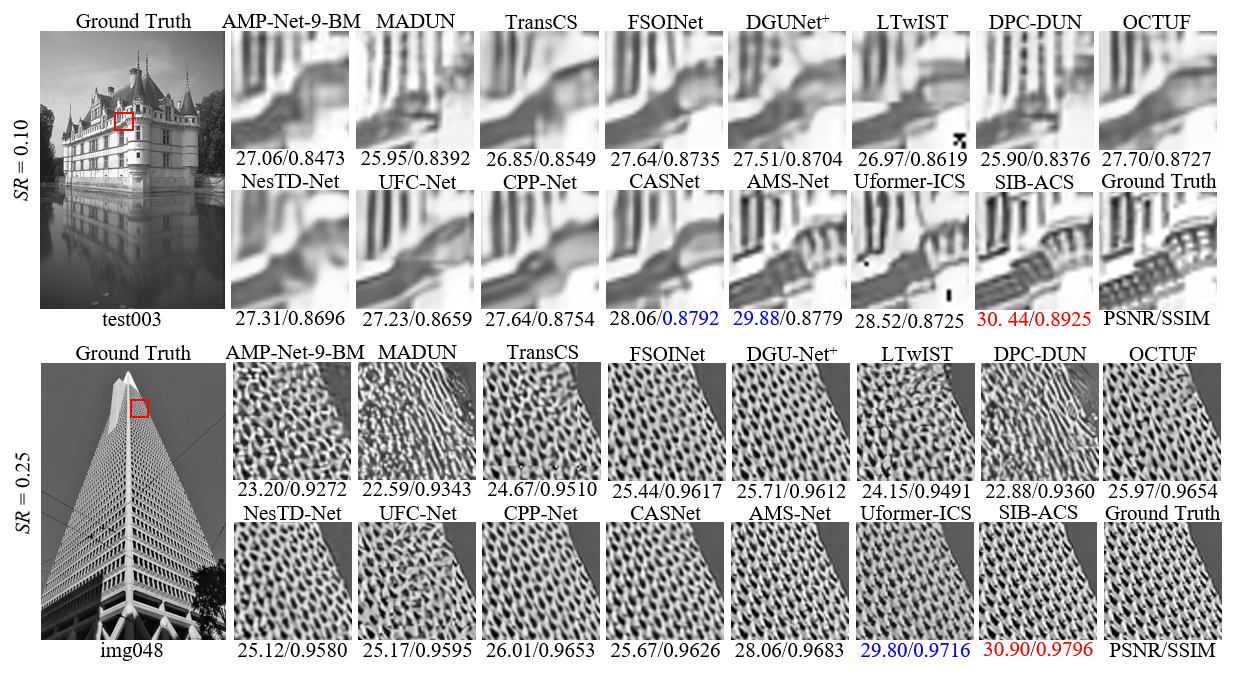}
	\caption{Visual comparisons of reconstructed image on test003 from BSD68~\cite{BSD68} at the sampling ratio of 0.10 and imag048 from Urban100~\cite{Urban100} at the sampling ratio of 0.25. The best and second-best results are marked in red and blue colors, respectively.}
	\label{fig:3}
	\vspace{-0.5cm}
\end{figure*}

\section{Experiments}
\label{sec:experiments}

\subsection{Experimental Settings}
In the SIB-ACS, we set the block size $B$ for the image to 32, and the initial sampling rate $SR_{\mathrm{init}}$ is defined to 0.02. Empirically, we determine the innovation sampling rate $SR_{\mathrm{IS}}$ as $\frac{SR-SR_{\mathrm{init}}}{2S}$, set the number of AS stages $S$ to 4~\cite{MGSA,Adacs}, and set $\mu$ to 0.1~\cite{MGSA}. The number of channels between iterative phases in the PCCD-Net is set to 32~\cite{OCTUF} and the dimension of the compressed domain is defined to 4. The number of phases $K$ in the PCCD-Net is set to 24. More implementation details and supplementary experiments can be found in the \textit{\textbf{Supplementary Material}}.

\subsection{Comparison with the state-of-the-arts}
We perform multiple qualitative comparison experiments between the proposed SIB-ACS and the recent state-of-the-art CS models~\cite{AMP-Net,MADUN,TransCS,FSOINET,DGUNet,LTwSTA,DPC-DUN,OCTUF,NesTD-Net,UFC,CPP,CASNet,AMS-Net,Uformer-ICS} at sampling rates of 0.10, 0.25, 0.30, 0.40, and 0.50, as shown in Tab.~\ref{tab:1}. The results show that the proposed SIB-ACS significantly surpasses existing advanced models in terms of image fidelity. The average PSNR(dB)/SSIM of the proposed SIB-ACS exceeds the second-best model on the BSD68 and Urban100 datasets by 1.10(3.17$\%$)/0.0101(1.10$\%$) and 1.48(4.25$\%$)/0.0028(0.30$\%$) respectively, evidencing the high-fidelity scene sensing capabilities of SIB-ACS.

Furthermore, we conduct visual comparisons of the proposed SIB-ACS with the recent state-of-the-art models on the BSD68~\cite{BSD68} and Urban100~\cite{Urban100} datasets at sampling rates of 0.10 and 0.25, as depicted in Fig.~\ref{fig:3}. As can be observed from Fig.~\ref{fig:3}, under various test dataset and sampling rate conditions, the proposed SIB-ACS is capable of more effectively identifying difficult-to-reconstruct areas in images and adaptively allocating sampling, which results in superior visual outcomes in the reconstruction of complex image areas, thereby enhancing the overall visual quality of the image.

\subsection{Effectiveness of ASM}
\label{sec:43}
\begin{table}[ht]
	\vspace{-0.3cm}
	\caption{PSNR ($\mathrm{dB}$) and SSIM comparisons of Uniform Sampling (US) and Adaptive Sampling (AS). The best results is marked in bold.}
	%\small
	\centering
	\resizebox{1.0\linewidth}{!}{
		\begin{tabular}{c|c|c|c|c}
			\toprule[1pt]
			\bottomrule[0.5pt]
			\rule{0pt}{10pt}\multirow{2}{*}{\makecell[c]{Sampling \\ module}} &  \multicolumn{2}{c|}{BSD68}& \multicolumn{2}{c}{Urban100}\cr\cline{2-5} 
			& \rule{0pt}{10pt}$SR = 0.10$ & $SR = 0.25$ & $SR = 0.10$ & $SR = 0.25$ \cr
			\cline{1-5}
			\rule{0pt}{13pt}US & 28.38/0.8207 & 32.15/0.9172 & 28.69/0.8807 & 33.11/0.9463 \cr\cline{1-5}
			\rule{0pt}{13pt}AS & \textbf{29.54}/\textbf{0.8401} & \textbf{34.35}/\textbf{0.9312} & \textbf{29.70}/\textbf{0.8850} & \textbf{35.15}/\textbf{0.9516} \cr
			\toprule[0.5pt]
			\toprule[1pt]
	\end{tabular}}
	\label{tab:2}
	\vspace{-0.3cm}
\end{table}
\begin{figure}[ht]
	\vspace{-0.3cm}
	\centering
	\setlength{\abovecaptionskip}{0.cm}
	\includegraphics[width=\columnwidth]{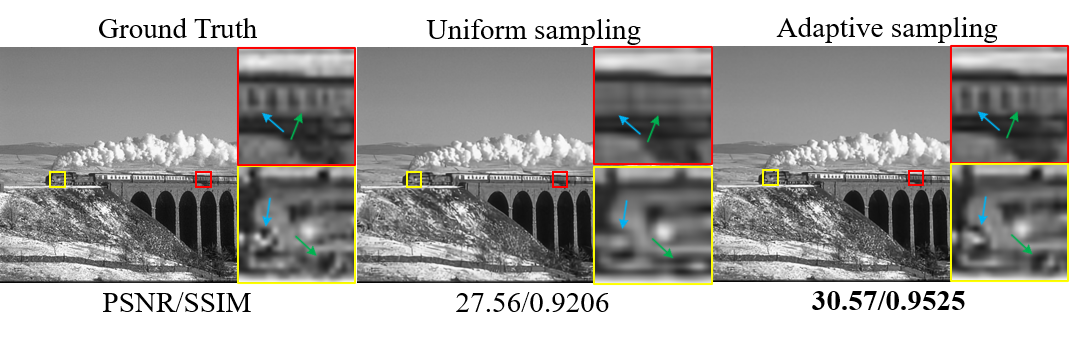}
	\caption{Visual comparisons of Uniform Sampling (US) and Adaptive Sampling (AS) on test033 from BSD68 at the sampling ratio of 0.25.}
	\label{fig:4}
	\vspace{-0.3cm}
\end{figure}

\noindent\textbf{Effectiveness of Adaptive Sampling}. We conduct comparative experiments between adaptive sampling (AS) and uniform sampling (US) to evaluate the effectiveness of AS, as presented in Tab.~\ref{tab:2}. US performs the same number of samples for each image block, while the proposed AS method adjusts the number of samples per image block based on the block's innovation. Both types of sampling use the same reconstruction network for image reconstruction. The results in Tab.~\ref{tab:2} show that compared to US, AS significantly improves the overall sensing ability of the image. Furthermore, we perform visual comparisons of the reconstruction results, as illustrated in Fig.~\ref{fig:4}. The visual quality of the reconstructed images suggest that, compared to US, the AS method can enhance sensing in areas with complex details, thereby improving the overall quality of image reconstruction.
\begin{table}[ht]
	\caption{PSNR ($\mathrm{dB}$) and SSIM comparisons of different Adaptive Sampling (AS) methods. The best results is marked in bold.}
	%\small
	\centering
	\resizebox{1.0\linewidth}{!}{
		\begin{tabular}{c|c|c|c|c}
			\toprule[1pt]
			\bottomrule[0.5pt]
			\rule{0pt}{10pt}\multirow{2}{*}{AS methods} & \multicolumn{2}{c|}{BSD68}& \multicolumn{2}{c}{Urban100}\cr\cline{2-5} 
			& \rule{0pt}{10pt}$SR = 0.10$ & $SR = 0.25$ & $SR = 0.10$ & $SR = 0.25$ \cr
			\cline{1-5}
			\rule{0pt}{13pt}Measurement Error  & 28.47/0.8124 & 32.63/0.9119 & 28.95/0.8732 & 34.16/0.9444 \cr\cline{1-5}
			\rule{0pt}{13pt}Saliency & 28.50/0.8169 & 32.59/0.9141 & 28.40/0.8715 & 33.71/0.9448 \cr\cline{1-5}
			\rule{0pt}{13pt}Sampling Innovation & \textbf{29.54}/\textbf{0.8401} & \textbf{34.35}/\textbf{0.9312} & \textbf{29.70}/\textbf{0.8850} & \textbf{35.15}/\textbf{0.9516} \cr
			\toprule[0.5pt]
			\toprule[1pt]
	\end{tabular}}
	\label{tab:3}
	\vspace{-0.3cm}
\end{table}
\begin{figure}[ht]
	%\vspace{-0.3cm}
	\centering
	\setlength{\abovecaptionskip}{0.cm}
	\includegraphics[width=\columnwidth]{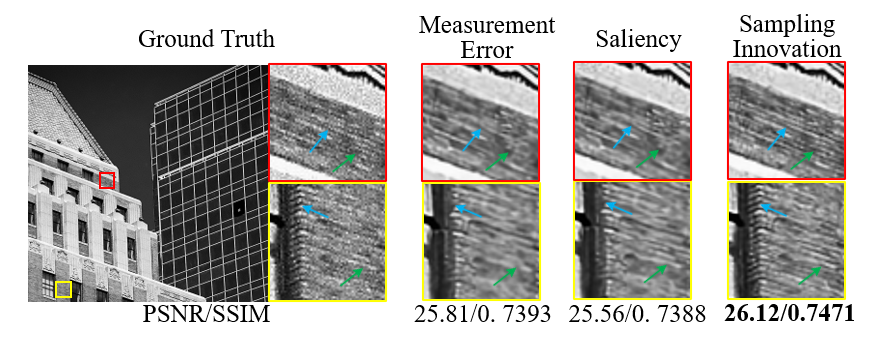}
	\caption{Visual comparisons of reconstructed image on img063 from Urban100~\cite{Urban100} at the sampling ratio of 0.25 using different AS methods.}
	\label{fig:9}
	\vspace{-0.5cm}
\end{figure}
\begin{figure}[ht]
	\vspace{-0.3cm}
	\centering
	\setlength{\abovecaptionskip}{0.cm}
	\includegraphics[width=\columnwidth]{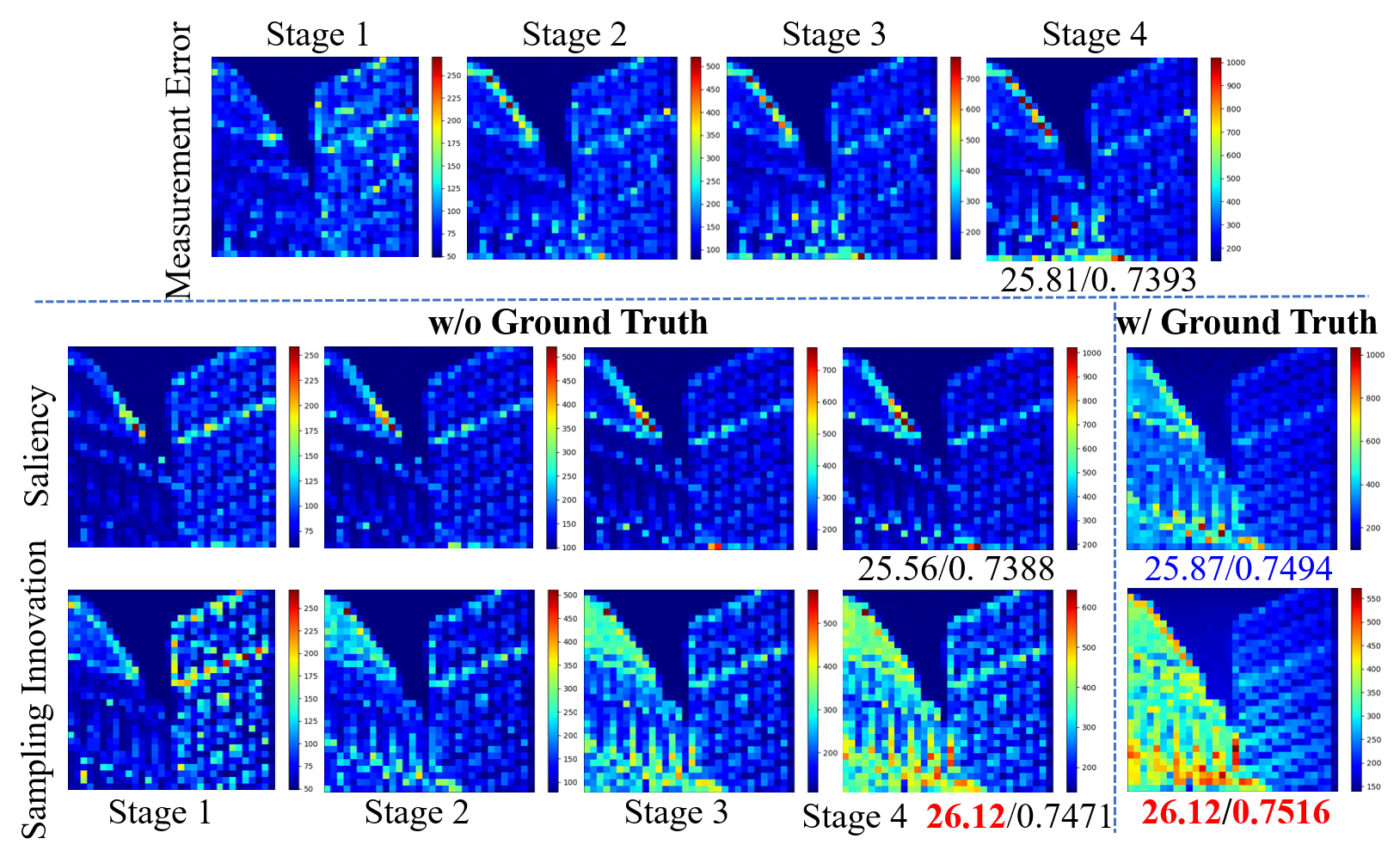}
	\caption{Visualization of sampling distribution at each stage in a multi-stage AS framework using different AS methods.}
	\label{fig:5}
	\vspace{-0.7cm}
\end{figure}

\noindent\textbf{Effectiveness of Innovation Criterion}. We compare the reconstruction results and the ASA results at each stage of the innovation-guided ACS framework with those of the measurement error~\cite{Adacs,RACSNet} and saliency~\cite{Uformer-ICS,CASNet,Sa}, as depicted in Tab.~\ref{tab:3} and Figs.~\ref{fig:9} and~\ref{fig:5} respectively. Table.~\ref{tab:3} and Fig.~\ref{fig:9} indicate that the innovation-guided multi-stage AS framework yields the best image reconstruction results. As illustrated in Fig.~\ref{fig:5}, the saliency-guided multi-stage ACS framework exhibits a positive feedback characteristic, where areas with high initial reconstruction saliency tend to accumulate samples, making it challenging to correct the ASA errors in areas that have high saliency in the original image but low initial reconstruction quality. The measurement error-guided multi-stage ACS framework has a negative feedback characteristic, capable of progressively correcting ASA errors. However, due to the influence of error clamping~\cite{Adacs,RACSNet}, the measurement error-guided model still exhibits non-negligible ASA errors in the end. Compared to the image-domain-based saliency ASA method, the innovation-guided ASA method exhibits negative feedback characteristics. Because innovation utilizes sampling increment information, a relative measure, which can reflect which image blocks' principal components have been restored as AS progresses and guides the ASA towards other image blocks that require more sampling, thereby progressively correcting the ASA. Simultaneously, compared to measurement error, because innovation is based on reconstructed image metrics, which contain richer image data information, the innovation-guided ASA possesses higher accuracy. In summary, the proposed innovation criterion is effective for guiding high-accuracy ASA.

\subsection{Effectiveness of PCCD-Net}
\label{sec:44}
\begin{table}[ht]
	\vspace{-0.3cm}
	\caption{Ablation experiments of PCPGD and CDPGD. The PC-Net and CD-Net are derived from the PCCD-Net by respectively omitting the CDPGD and PCPGD paths. The best results is marked in bold.}
	%\small
	\centering
	\resizebox{1.0\linewidth}{!}{
		\begin{tabular}{c|c|c|c|c|c|c}
			\toprule[1pt]
			\bottomrule[0.5pt]
			\rule{0pt}{10pt}\multirow{2}{*}{Models} & \multirow{2}{*}{PCPGD} & \multirow{2}{*}{CDPGD} & \multicolumn{2}{c|}{BSD68} & \multicolumn{2}{c}{Urban100}\cr\cline{4-7} 
			& & & \rule{0pt}{10pt}$SR = 0.10$ & $SR = 0.25$ & $SR = 0.10$ & $SR = 0.25$ \cr
			\cline{1-7}
			\rule{0pt}{13pt}PC-Net 
			& $\surd$ & $\times$ & 28.32/0.8197 & 32.11/0.9169 & 28.37/0.8759 & 32.94/0.9452  \cr
			\cline{1-7}
			\rule{0pt}{13pt}CD-Net
			& $\times$ & $\surd$ & 28.22/0.8171 & 32.10/0.9166 & 28.10/0.8710 & 32.83/0.9438 \cr
			\cline{1-7}
			\rule{0pt}{13pt}PCCD-Net
			& $\surd$ & $\surd$ & \textbf{28.38}/\textbf{0.8207} & \textbf{32.15}/\textbf{0.9172} & \textbf{28.69}/\textbf{0.8807} & \textbf{33.11}/\textbf{0.9463}  \cr
			\toprule[0.5pt]
			\toprule[1pt]
	\end{tabular}}
	\label{tab:4}
	\vspace{-0.3cm}
\end{table}
\begin{figure}[ht]
	\vspace{-0.3cm}
	\centering
	\setlength{\abovecaptionskip}{0.cm}
	\includegraphics[width=0.8\columnwidth]{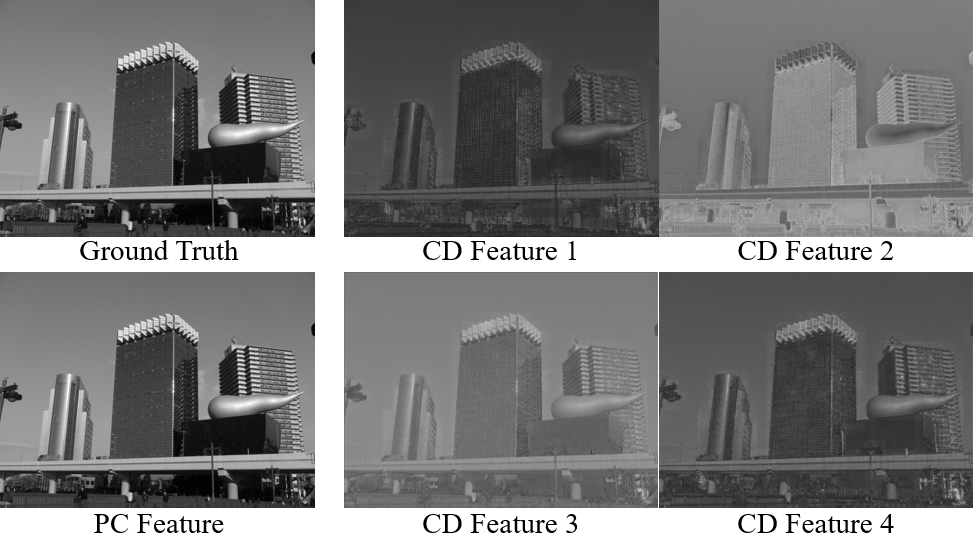}
	\caption{Visualization of principal component (PC) image in PCPGD and compressed domain (CD) features in CDPGD.}
	\label{fig:6}
	\vspace{-0.3cm}
\end{figure}

\noindent\textbf{Ablation of PCPGD and CDPGD}. To verify the contributions of PCPGD and CDPGD to the PCCD-Net, we conduct corresponding ablation experiments, as depicted in Tab.~\ref{tab:4}. The results indicate that the removal of CDPGD and PCPGD from PCCD-Net leads to varying degrees of decline in reconstruction performance, demonstrating the effectiveness of CDPGD and PCPGD. To ensure that the performance improvement is not merely due to the larger model parameters in PCCD-Net, more detailed convergence verification is provided in the \textit{\textbf{Supplementary Material}}. Additionally, we visualize the features undergoing the PGD operation in CDPGD and PCPGD, as illustrated in Fig.~\ref{fig:6}. The visualization results confirm that PCPGD executes the PGD of the principal component image, while CDPGD carries out the PGD of the supplementary features in the compressed domain, ultimately achieving high-quality reconstruction of image features.
\begin{table}[ht]
	\vspace{-0.3cm}
	\caption{Ablation experiments of compressed domain (CD) dimension. The best results is marked in bold.}
	%\small
	\centering
	\resizebox{1.0\linewidth}{!}{
		\begin{tabular}{c|c|c|c|c|c|c}
			\toprule[1pt]
			\bottomrule[0.5pt]
			\rule{0pt}{10pt}\multirow{2}{*}{\makecell[c]{CD \\ Dimension}} & \multicolumn{2}{c|}{BSD68}& \multicolumn{2}{c|}{Urabn100} & \multicolumn{2}{c}{MACs (T)} \cr\cline{2-7} 
			& \rule{0pt}{10pt} $SR = 0.10$ & $SR = 0.25$ & $SR = 0.10$ & $SR = 0.25$ & $SR = 0.25$ & $SR = 1.00$ \cr
			\cline{1-7}
			\rule{0pt}{13pt}2  & 28.30/0.8191 & 32.12/0.9169 & 28.51/0.8783 & 32.93/0.9451 & 1.4 & 5.2 \cr\cline{1-7}
			\rule{0pt}{13pt}4  & \textbf{28.38}/\textbf{0.8207} & 32.15/0.9172 & \textbf{28.69}/\textbf{0.8807} & 33.11/0.9463& 2.2& 8.5 \cr\cline{1-7}
			\rule{0pt}{13pt}8 & \textbf{28.38}/0.8205 & \textbf{32.19}/\textbf{0.9178} & 28.65/0.8798 & 33.14/0.9467& 3.9 & 15.1  \cr\cline{1-7}
			\rule{0pt}{13pt}16 & 28.33/0.8202 & 32.16/0.9175 & 28.54/0.8789 & 33.16/\textbf{0.9471} & 7.2 & 28.3 \cr\cline{1-7}
			\rule{0pt}{13pt}32 & 28.32/0.8198 & \textbf{32.19}/\textbf{0.9178} & 28.51/0.8788 & \textbf{33.19}/\textbf{0.9471}& 13.8 & 54.7 \cr
			\toprule[0.5pt]
			\toprule[1pt]
	\end{tabular}}
	\label{tab:5}
	\vspace{-0.3cm}
\end{table}

\noindent\textbf{Ablation of Compressed Domain (CD) Dimension}. We investigate the performance of PCCD-Net with varying CD dimensions. When the CD dimensions are set to 2, 4, 8, 16, and 32 respectively, the reconstruction quality and Multiply-Accumulate Operations (MACs) of PCCD-Net are presented in Tab.~\ref{tab:5}. The results indicate an proportional  increase in the model's computational load corresponding to the rise in feature and sampling matrix dimensions during the gradient descent operation. In ACS scenarios, which are typically characterized by a high upper limit of the block image sampling rate, careful design is required to manage the computational load introduced by the gradient descent operation. We observe an oscillation in the image reconstruction performance when the CD dimension exceeds 4. Considering both the reconstruction quality and computational load, we choose a CD dimension of 4.

\subsection{Complexity Analysis}
\begin{table}[ht]
	\vspace{-0.3cm}
	\caption{Comparisons of the model average PSNR, parameter size, and running time. The best results is marked in bold.}
	\centering
	\resizebox{1.0\linewidth}{!}{
		\begin{tabular}{c|ccccc|cc}
			\toprule[1pt]
			\bottomrule[0.5pt]
			\rule{0pt}{10pt}Models  & OCTUF& NesTD-Net& UFC-Net& CPP-Net & CASNet & PCCD-Net & SIB-ACS\cr
			\cline{1-8} 
			\rule{0pt}{13pt}PSNR (dB)  & 30.31 & 30.32& 30.01 & 30.63 & 30.09 & 30.58 & \textbf{32.19} \cr
			\rule{0pt}{13pt}Params (M)  & \textbf{0.29}& 5.93& 1.90 & 12.47 & 15.85 & 2.32 & 4.73 \cr
			\rule{0pt}{13pt}Time (s)&0.061 & 0.196& 0.166& 0.193 &0.106 &\textbf{0.060}&0.160 \cr
			\toprule[0.5pt]
			\toprule[1pt]
	\end{tabular}}
	\label{tab:6}
	\vspace{-0.3cm}
\end{table}
\noindent Computational complexity is a crucial reference indicator for model application. We evaluate the comprehensive performance of the proposed ACS model SIB-ACS and UCS model PCCD-Net by comparing the reconstruction quality, model parameter size, and runtime with state-of-the-art methods, as depicted in Tab.~\ref{tab:6}. The PSNR in Tab.~\ref{tab:5} represents the average PSNR of the BSD68~\cite{BSD68} and Urban100~\cite{Urban100} datasets under sampling rates of 0.10 and 0.25. We measure the runtime by obtaining sampling values at a sampling rate of 0.25 on 256$\times$256 images and then reconstructing the images. The results indicate that the runtime of the proposed PCCD-Net is comparable to that of the lightweight OCTUF~\cite{OCTUF}, but the image reconstruction quality of PCCD-Net surpasses that of OCTUF. The image reconstruction quality of the proposed PCCD-Net is second only to CPP-Net~\cite{CPP}, but the model parameters and runtime of PCCD-Net are significantly less than those of CPP-Net. Furthermore, under conditions of comparable model parameter size and runtime, the proposed ACS model SIB-ACS achieves the highest fidelity in image reconstruction.

\section{Conclusion}
\label{sec:conclusion}
In this paper, we propose a sampling innovation-based adaptive compressive sensing (SIB-ACS) framework that leverages image increment information for negative feedback to facilitate adaptive sampling allocation (ASA), achieving high-fidelity image reconstruction. The innovation criterion is proposed to adjudicate the ASA, and an innovation-guided multi-stage ASA framework is established to correct the ASA stage by stage. In addition, the proposed Principal Component Compressed Domain Network (PCCD-Net) ensures high-fidelity image reconstruction while controlling the computational burden of the gradient descent in the ACS scenario. Extensive experiments have confirmed the effectiveness of the proposed SIB-ACS and its capability to reconstruct high-fidelity scenes.

{
    \small
    \bibliographystyle{ieeenat_fullname}
    \bibliography{main}
}

\end{document}